\documentclass[runningheads]{llncs}
\usepackage{graphicx}

\usepackage{tikz}
\usepackage{comment}
\usepackage{amsmath,amssymb} 
\usepackage{color}
\usepackage[pagebackref=true,breaklinks=true,colorlinks,bookmarks=false]{hyperref}

\usepackage[accsupp]{axessibility}  

\usepackage[width=122mm,left=12mm,paperwidth=146mm,height=193mm,top=12mm,paperheight=217mm]{geometry}

\usepackage{subcaption}
\usepackage{nccmath}
\usepackage{here}

\usepackage{bbold}
\usepackage{multirow}
\usepackage{bm}
\usepackage{arydshln}
\usepackage{xspace}

\makeatletter
\newcommand{\figcaption}[1]{\def\@captype{figure}\caption{#1}}
\newcommand{\tblcaption}[1]{\def\@captype{table}\caption{#1}}
\makeatother

\DeclareRobustCommand\red{\textcolor{red}}
\DeclareRobustCommand\blue{\textcolor{blue}}

\def\tbf{\textbf}

\def\bf{\textbf}
\def\it{\textit}
\def\bs{\boldsymbol}

\makeatletter
\DeclareRobustCommand\onedot{\futurelet\@let@token\@onedot}
\def\@onedot{\ifx\@let@token.\else.\null\fi\xspace}
\def\eg{\emph{e.g}\onedot} 
\def\ie{\emph{i.e}\onedot} 
 
 \def\vs{\emph{vs}\onedot}
 
\def\etal{\emph{et al}\onedot}
\def\aka{\emph{a.k.a}\onedot}
\makeatother

\def\tabledho{
\begin{table}[!t]
\footnotesize{
\begin{center}
\caption{\tbf{DexYCB~\cite{chao:cvpr21} → HO3D~\cite{hampali:cvpr20}}. 
We report PCK (\%) and MPE (px) for hand keypoint regression and IoU (\%) for hand segmentation. 
Each score format of $val~/~test$ indicates the validation and test scores.
Red and blue letters indicate the best and second best values.
}
\label{tbl:d2ho}
\scalebox{0.88}[0.9]{
\begin{tabular}{l|cccc}
\multirow{2}{*}{Method} & \multicolumn{2}{c}{2D Pose}                                         & \multicolumn{1}{c|}{Seg}         & 2D Pose + Seg                  \\
                        & \multicolumn{1}{c}{PCK~↑~(\%)}  & \multicolumn{1}{c}{MPE~↓~(px)}     & \multicolumn{1}{c|}{IoU~↑~(\%)}   & \multicolumn{1}{c}{Avg.~↑~(\%)} \\ \hline
Source only             & $42.8 / 33.5$                     & $15.39 / 19.32$                     & \multicolumn{1}{c|}{$57.9 / 49.1$} & $50.3 / 41.3$                    \\ 
\hdashline[2pt/2pt]
DANN~\cite{ganin:icml15}& $49.0 / 46.8$                     & $12.39 / 13.39$                     & \multicolumn{1}{c|}{$52.8 / 54.7$} & $50.9 / 50.8$                    \\
RegDA~\cite{jiang:cvpr21}& $48.8 / 48.2$	                  & $12.50 / 12.64$                     & \multicolumn{1}{c|}{$55.7 / 55.3$} & $52.2 / 51.7$\\
GAC                      & $47.6 / 47.4$                     & $12.47 / 12.54$                     & \multicolumn{1}{c|}{$58.0 / 56.9$} & $52.8 / 52.2$                    \\
GAC + UMA~\cite{cai:cvpr20} & $47.1 / 45.3$ & $12.97 / 13.51$ & \multicolumn{1}{c|}{$58.0 / 55.0$} & $52.5 / 50.2$ \\
GAC + CPL~\cite{ohkawa:access21}& $48.1 / 48.1$                     & $12.74 / 12.61$                     & \multicolumn{1}{c|}{$57.2 / 55.6$} & $52.7 / 51.8$                    \\
GAC + MT~\cite{tarvainen:iclr17}& $45.5 / 44.4$                     & $13.65 / 14.05$                     & \multicolumn{1}{c|}{$54.8 / 52.3$} & $50.2 / 48.3$                    \\ 
\hdashline[2pt/2pt]
GAC-Distill (Ours)      & \multicolumn{1}{c}{$\blue{49.9} / \blue{50.4}$} & \multicolumn{1}{c}{$\blue{11.98} / \blue{11.51}$} & \multicolumn{1}{c|}{$\blue{60.7} / \red{60.6}$} & $\blue{55.3} / \blue{55.5}$                    \\
C-GAC (Ours-Full)        & $\red{50.3} / \red{51.1}$ & $\red{11.89} / \red{11.22}$ &  \multicolumn{1}{c|}{$\red{60.9} / \blue{60.3}$} &  $\red{55.6} / \red{55.7}$ \\ \hline
Target only             & $55.1 / 58.6$                     & $11.00 / 9.29$                      & \multicolumn{1}{c|}{$68.2 / 66.1$} & $61.7 / 62.4$   
\end{tabular}
}
\end{center}
}
\end{table}
}

\def\tabledhaf{
\begin{table}[t]
  \caption{\tbf{DexYCB~\cite{chao:cvpr21} → \{HanCo~\cite{zimmermann:arxiv21},  FPHA~\cite{hernando:cvpr18}\}}. 
    We report PCK (\%) and MPE (px) for hand keypoint regression and IoU (\%) for hand segmentation. 
    We show the validation and test results on HanCo and the validation results on FPHA.
    Red and blue letters indicate the best and second best values.
  }
  \label{tbl:d2haf}
  \begin{minipage}[t!]{.65\textwidth}
   \footnotesize{
    \begin{center}
    \scalebox{0.88}[0.9]{
    \begin{tabular}{l|cccc}
    \multirow{3}{*}{Method} & \multicolumn{4}{|c}{DexYCB → HanCo} \\ \cdashline{2-5} 
                            & \multicolumn{2}{c}{2D Pose}                                         & \multicolumn{1}{c|}{Seg}         &      \multicolumn{1}{c}{}            \\
                            & \multicolumn{1}{c}{PCK~↑~(\%)}  & \multicolumn{1}{c}{MPE~↓~(px)}     & \multicolumn{1}{c|}{IoU~↑~(\%)}   & \multicolumn{1}{c}{Avg.~↑~(\%)} \\ \hline
    Source only             & $26.0 / 27.3$ & $21.82 / 21.48$ &  \multicolumn{1}{c|}{$41.8 / 41.4$} & $33.9 / 34.3$ \\
    \hdashline[2pt/2pt]
    DANN~\cite{ganin:icml15}& $32.3 / 33.0$ & $19.99 / 19.82$ & \multicolumn{1}{c|}{$56.3 / 56.9$} & $44.3 / 45.0$ \\
    RegDA~\cite{jiang:cvpr21}& $33.0 / 33.6$ & $19.51 / 19.44$ & \multicolumn{1}{c|}{$57.8 / 58.4$} & $45.4 / 46.0$ \\
    GAC                      & $36.6 / 37.1$ & $16.63 / 16.59$ & \multicolumn{1}{c|}{$\blue{58.1} / \red{58.8}$} & $47.4 / 47.9$ \\
    GAC + UMA~\cite{cai:cvpr20}& $35.1 / 35.6$ & $17.51 / 17.48$ &  \multicolumn{1}{c|}{$57.1 / 57.7$} & $46.1 / 46.6$ \\
    GAC + CPL~\cite{ohkawa:access21}& $32.7 / 33.5$ & $19.85 / 19.62$ & \multicolumn{1}{c|}{$55.8 / 56.4$} & $44.2 / 45.0$ \\
    GAC + MT~\cite{tarvainen:iclr17}& $33.2 / 33.8$ & $18.93 / 18.83$ & \multicolumn{1}{c|}{$54.3 / 55.1$} & $43.8 / 44.4$ \\  
    \hdashline[2pt/2pt]
    GAC-Distill (Ours) & $\blue{38.8} / \blue{39.5}$ & $\blue{16.06} / \blue{15.97}$ &  \multicolumn{1}{c|}{$57.5 / 57.7$} & $\blue{48.1} / \blue{48.6}$ \\
    C-GAC (Ours-Full)  & $\red{39.2} / \red{39.9}$ & $\red{15.83} / \red{15.74}$ &  \multicolumn{1}{c|}{$\red{58.2} / \blue{58.6}$} & $\red{48.7} / \red{49.2}$  \\ \hline
    Target only        & $76.8 / 77.3$ & $4.91 / 4.80$   & \multicolumn{1}{c|}{$75.9 / 76.1$} & $76.3 / 76.7$ \\
    \end{tabular}
    }
    \end{center}
    }
  \end{minipage}
  \hfill
  \begin{minipage}[t!]{.3\textwidth}
   \footnotesize{
    \begin{center}
    \scalebox{0.88}[0.9]{
    \begin{tabular}{||cccc}
    \multicolumn{4}{||c}{DexYCB → FPHA} \\ \cdashline{1-4}
    \multicolumn{2}{||c}{2D Pose}                                         & \multicolumn{1}{c|}{Seg}         &                  \\
                            \multicolumn{1}{||c}{PCK}  & \multicolumn{1}{c}{MPE}     & \multicolumn{1}{c|}{IoU}   & \multicolumn{1}{c}{Avg.} \\ \hline
    14.0 & 31.32 & \multicolumn{1}{c|}{24.8} & 19.4 \\
    \hdashline[2pt/2pt]
    24.4 & 25.79 & \multicolumn{1}{c|}{28.4} & 26.4 \\
    23.7 & 24.27 & \multicolumn{1}{c|}{\red{41.7}} & 32.7 \\
    \red{37.2} & 17.02 & \multicolumn{1}{c|}{33.3} & 35.3 \\
    \blue{36.8} & 17.29 &  \multicolumn{1}{c|}{\blue{39.2}} & \red{38.0} \\
    25.7 & 24.99 & \multicolumn{1}{c|}{32.7} & 29.2 \\ 
    31.3 & 20.81 & \multicolumn{1}{c|}{38.4} & 34.9 \\ 
    \hdashline[2pt/2pt]
    \blue{36.8} & \blue{15.99} & \multicolumn{1}{c|}{35.5} & 36.1 \\
    \red{37.2} & \red{15.36} &  \multicolumn{1}{c|}{37.7} & \blue{37.4} \\ \hline
    63.3 &  8.11 &  \multicolumn{1}{c|}{-} & - \\
    \end{tabular}
    }
    \end{center}
    }
  \end{minipage}
\end{table}
}

\begin{document}
\pagestyle{headings}
\mainmatter

\title{Domain Adaptive Hand Keypoint and Pixel Localization in the Wild}


\titlerunning{Domain Adaptive Hand Keypoint and Pixel Localization in the Wild}
\author{Takehiko Ohkawa\inst{1,2}, Yu-Jhe Li\inst{2}, Qichen Fu\inst{2}, Ryosuke Furuta\inst{1}, Kris~M.~Kitani\inst{2}, and Yoichi Sato\inst{1}}
\authorrunning{T. Ohkawa et al.}
\institute{The University of Tokyo, Tokyo, Japan \and
Carnegie Mellon University, PA, USA \\
\email{\{ohkawa-t,furuta,ysato\}@iis.u-tokyo.ac.jp, \{yujheli,qichenf,kkitani\}@cs.cmu.edu}\\
Project: \href{https://tkhkaeio.github.io/projects/22-hand-ps-da/}{https://tkhkaeio.github.io/projects/22-hand-ps-da/}
}

\maketitle
\begin{abstract}
We aim to improve the performance of regressing hand keypoints and segmenting pixel-level hand masks under new imaging conditions (\eg, outdoors) when we only have labeled images taken under very different conditions (\eg, indoors). 
In the real world, it is important that the model trained for both tasks works under various imaging conditions.
However, their variation covered by existing labeled hand datasets is limited.
Thus, it is necessary to adapt the model trained on the labeled images (source) to unlabeled images (target) with unseen imaging conditions.
While self-training domain adaptation methods (\ie, learning from the unlabeled target images in a self-supervised manner) have been developed for both tasks, their training may degrade performance when the predictions on the target images are noisy.
To avoid this, it is crucial to assign a low importance (confidence) weight to the noisy predictions during self-training.
In this paper, we propose to utilize the divergence of two predictions to estimate the confidence of the target image for both tasks.
These predictions are given from two separate networks, and their divergence helps identify the noisy predictions.
To integrate our proposed confidence estimation into self-training, we propose a teacher-student framework where the two networks (teachers) provide supervision to a network (student) for self-training, and the teachers are learned from the student by knowledge distillation.
Our experiments show its superiority over state-of-the-art methods in adaptation settings with different lighting, grasping objects, backgrounds, and camera viewpoints.
Our method improves by $4\%$ the multi-task score on HO3D compared to the latest adversarial adaptation method.
We also validate our method on Ego4D, egocentric videos with rapid changes in imaging conditions outdoors.
\end{abstract}

\section{Introduction}\label{sec:introduction}

\begin{figure*}[t]
\centering
\includegraphics[width=0.95\hsize]{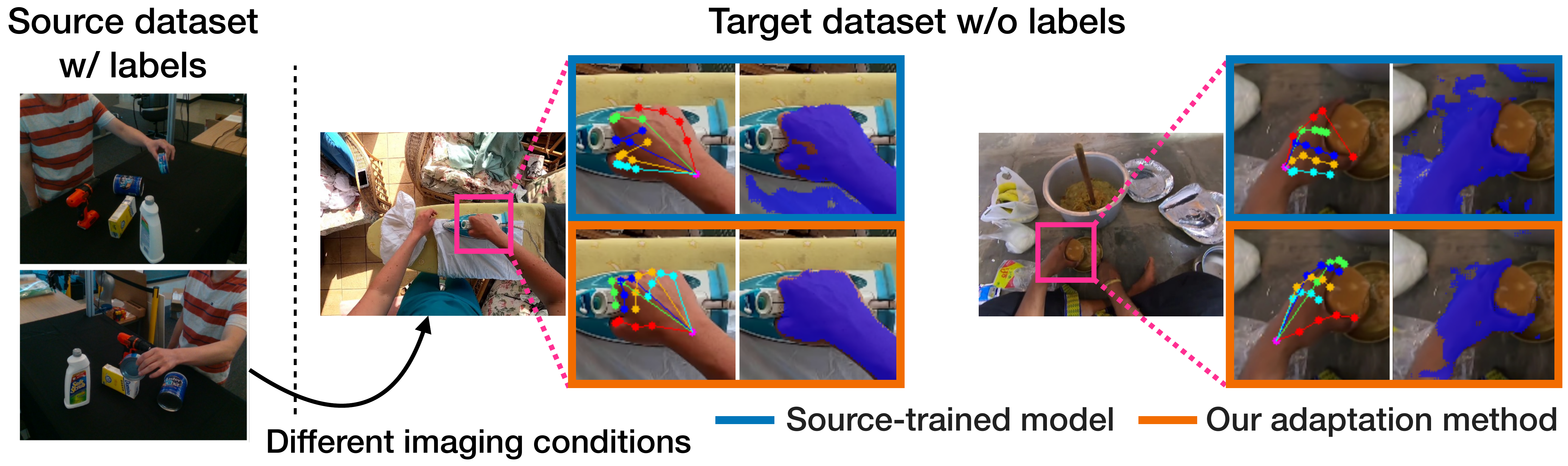}
\caption{
We aim to adapt the model of localizing hand keypoints and pixel-level hand masks to new imaging conditions without annotation.
}
\label{fig:teaser}
\end{figure*}

In the real world, hand keypoint regression and hand segmentation are considered important to work under broad imaging conditions for various computer vision applications, such as egocentric video understanding~\cite{grauman:cvpr22,damen:ijcv21}, hand-object interaction analysis~\cite{cao:iccv21,fu:arxiv21}, AR/VR~\cite{liang:acmmm15,wu:vcir20}, and assistive technology~\cite{likitlersuang:jner21,lee:wacv20}.
For building models for both tasks, several labeled hand datasets have been proposed in laboratory settings, such as multi-camera studios~\cite{joo:cvpr15,moon:eccv20,chao:cvpr21,zimmermann:iccv19} and attaching sensors to hands~\cite{hernando:cvpr18,yuan:cvpr17,glauser:tog19}.
However, their imaging conditions do not adequately cover real-world imaging conditions~\cite{ohkawa:arxiv22}, consisting of various lighting, hand-held objects, backgrounds, and camera viewpoints.
In addition, the annotation of keypoints and pixel-level masks are not always available in real-world environments because they are labor-intensive to acquire.
As shown in Fig.~\ref{fig:teaser}, when localizing hand keypoints and pixels in real-world egocentric videos~\cite{grauman:cvpr22} (\eg, outdoors), we may only have access to a hand dataset~\cite{chao:cvpr21} taken under completely different imaging conditions (\eg, indoors).
Given these limitations, we need methods that can robustly adapt the models trained on the available labeled images (source) to unlabeled images (target) with new imaging conditions.

To enable such adaptation, the approach of self-training domain adaptation has been developed for both tasks.
This approach aims to learn unlabeled target images by optimizing a self-supervised task, which exhibits effectiveness in various domain adaptation tasks~\cite{chen:nips11,deng:cvpr21,vu:cvpr19,zhou:eccv18,cai:cvpr20}.
For keypoint estimation, consistency training, a method that regularizes keypoint predictions to be consistent under geometric transformations, has been proposed~\cite{yang:iccv21,vasconcelos:icpr20,zhou:eccv18}.
As for hand segmentation, prior studies use pseudo-labeling~\cite{ohkawa:access21,cai:cvpr20}, which produces hard labels by thresholding a predicted class probability for updating a network.
However, these self-training methods for both tasks perform well only when the predictions are reasonably correct.
When the predictions become noisy due to the gap in imaging conditions, the trained network will cause over-fitting to the noisy predictions, resulting in poor performance in the target domain.

To avoid this, it is crucial to assign a low importance (confidence) weight to the loss of self-training with noisy predictions.
This confidence weighting can mitigate the distractions from the noisy predictions.
To this end, we propose self-training domain adaptation with confidence estimation for hand keypoint regression and hand segmentation.
Our proposed method consists of (i) confidence estimation based on the divergence of two networks' predictions and (ii) an update rule that integrates a training network for self-training and the two networks for confidence estimation.

To (i) estimate confidence, we utilize the predictions of two different networks.
While class probability can be used as the confidence in classification tasks, it is not trivial to obtain such a measure in keypoint regression.
Thus, we newly focus on the divergence of the two networks' predictions for each target image.
We design their networks to have an identical architecture but have different learning parameters.
We observe that when the divergence measure is high, the predictions of both networks are noisy and should be avoided in self-training.

To (ii) integrate the estimated confidence into self-training, inspired by the single-teacher-single-student update~\cite{tarvainen:iclr17,pham:cvpr21}, we develop mutual training with self-training based on consistency training for a training network (student) and distillation-based update for the two networks (teachers).
For training the student network, we build a unified self-training framework that can work favorably for the two tasks.
Motivated by supervised or weakly-supervised learning for jointly estimating both tasks~\cite{wang:tcsvt19,zhang:icassp20,goudie:fg17,neverova:cviu17,chen:access18}, we expect that jointly adapting both tasks will allow one task to provide useful cues to the other task even in the unlabeled target domain.
Specifically, we enforce the student network to generate consistent predictions for both tasks under geometric augmentation.
We weight the loss of the consistency training using the confidence estimated from the divergence of the teachers' predictions.
This can reduce the weight of the noisy predictions during the consistency training.
To learn the two teacher networks differently, we train the teachers independently from different mini-batches by knowledge distillation, which matches the teacher-student predictions in the output level.
This framework enables the teachers to update more carefully than the student and prevent over-fitting to the noisy predictions.
Such stable teachers provide reliable confidence estimation for the student's training.

In our experiments, we validate our proposed method in adaptation settings where lighting, grasping objects, backgrounds, camera viewpoints, etc., vary between labeled source images and unlabeled target images.
We use a large-scale hand dataset captured in a multi-camera system~\cite{chao:cvpr21} as the source dataset (see Fig.~\ref{fig:teaser}).
For the target dataset, we use HO3D~\cite{hampali:cvpr20} with different environments, HanCo~\cite{zimmermann:arxiv21} with multiple viewpoints and diverse backgrounds, and FPHA~\cite{hernando:cvpr18} with a novel first-person camera viewpoint.
We also apply our method to in-the-wild egocentric video Ego4D~\cite{grauman:cvpr22} (see Fig.~\ref{fig:teaser}), including diverse indoor and outdoor activities worldwide.
Our method improves the average score of the two tasks by $14.4\%$, $14.9\%$, and $18.0\%$ on HO3D, HanCo, and FPHA, respectively, compared to a unadapted baseline.
Our method further exhibits distinct improvements compared to the latest adversarial adaption method~\cite{jiang:cvpr21} and consistency training baselines with uncertainty estimation~\cite{cai:cvpr20}, confident instance selection~\cite{ohkawa:access21}, and the teacher-student scheme~\cite{tarvainen:iclr17}.
We finally confirm that our method also performs qualitatively well on the Ego4D videos.

Our contributions are summarized as follows:
\begin{itemize}
    \vspace{-5pt}
    \setlength{\parskip}{0pt}
    \setlength{\itemsep}{3pt}
    \item We propose a novel confidence estimation method based on the divergence of the predictions from two teacher networks for self-training domain adaptation of hand keypoint regression and hand segmentation.
    \item To integrate our proposed confidence estimation into self-training, we propose mutual training using knowledge distillation with a student network for self-training and two teacher networks for confidence estimation.
    \item Our proposed framework outperforms state-of-the-art methods under three adaptation settings across different imaging conditions. It also shows improved qualitative performance on in-the-wild egocentric videos.
\end{itemize}

\section{Related Work}\label{sec:relatedwork}

\tbf{Hand keypoint regression} is the task of regressing the positions of hand joint keypoints from a cropped hand image.
2D hand keypoint regression is trained by optimizing keypoint heatmaps~\cite{wei:cvpr16,newell:eccv16,zimmermann:iccv17} or directly predicting keypoint coordinates~\cite{santavas:arxiv20}.
The 2D keypoints are informative for estimating 3D hand poses~\cite{simon:cvpr17,yang:bmvc20,mueller:cvpr18,adnane:cvpr19}.
To build an accurate keypoint regressor, collecting massive hand keypoint annotations is required but laborious.
While early works annotate the keypoints manually from a single view~\cite{qian:cvpr14,sridhar:eccv16,mueller:iccv17}, recent studies have collected the annotation more densely and efficiently using synthetic hand models~\cite{hasson:cvpr19,mueller:cvpr18,zimmermann:iccv17,mueller:iccv17}, hand sensors~\cite{hernando:cvpr18,yuan:cvpr17,taheri:eccv20,glauser:tog19}, or multi-camera setups~\cite{joo:cvpr15,moon:eccv20,chao:cvpr21,brahmbhatt:eccv20,zimmermann:iccv19,hampali:cvpr20,lu:arxiv21}.
However, these methods suffer the gap in imaging conditions with real-world images in deployment~\cite{ohkawa:arxiv22}.
For instance, the synthetic hand models and hand sensors induce different lighting conditions from actual human hands.
The multi-camera setup lacks a variety of lighting, grasping objects, and backgrounds.
To tackle these problems, domain adaptation is a promising solution that can transfer the knowledge of the network trained on source data to unlabeled target data.
Jiang~\etal proposed an adversarial domain adaptation for human and hand keypoint regression, optimizing the discrepancy between regressors~\cite{jiang:cvpr21}.
Additionally, self-training adaptation methods have been studied in the keypoint regression of animals~\cite{cao:iccv19}, humans~\cite{vasconcelos:icpr20}, and objects~\cite{zhou:eccv18}.
Unlike these prior works, we incorporate confidence estimation into a self-training method based on consistency training for keypoint regression. 

\tbf{Hand segmentation} is the task of segmenting pixel-level hand masks in a given image.
CNN-based segmentation networks~\cite{urooj:cvpr18,garcia:sensors21,kim:bmvc20} are popularly used.
The task can be jointly trained with hand keypoint regression because detecting hand regions guides to improve keypoint localization~\cite{wang:tcsvt19,zhang:icassp20,goudie:fg17,neverova:cviu17,chen:access18}.
Since hand mask annotation is laborious as hand keypoint regression, a few domain adaptation methods with pseudo-labeling have been explored~\cite{cai:cvpr20,ohkawa:access21}.
To reduce the effect of highly noisy pseudo-labels in the target domain, Cai~\etal incorporate the uncertainty of pseudo-labels in model adaptation~\cite{cai:cvpr20}, and Ohkawa~\etal select confident pseudo-labels by the overlap of two predicted hand masks~\cite{ohkawa:access21}.
Unlike~\cite{cai:cvpr20}, we estimate the target confidence using two networks.
Instead of using the estimated confidence for instance selection~\cite{ohkawa:access21}, we assign the confidence to weight the loss of consistency training.

\tbf{Domain adaptation via self-training} aims to learn unlabeled target data in a self-supervised learning manner.
This approach can be divided into three categories.
(i) Pseudo-labeling~\cite{chen:nips11,saito:icml17,zou:eccv18,ohkawa:access21,cai:cvpr20} learns unlabeled data with hard labels assigned by confidence thresholding from the output of a network.
(ii) Entropy minimization~\cite{long:nips16,vu:cvpr19,prabhu:iccv21} regularizes the conditional entropy of unlabeled data and increases the confidence of class probability. 
(iii) Consistency regularization~\cite{xie:nips20,chen:cvpr19,fu:cvpr19} enforces regularization so that the prediction on unlabeled data is invariant under data perturbation.
We choose to leverage this consistency-based method for our task because
it works for various tasks~\cite{kyriazi:cvpr21,liu:iclr21,ohkawa:icpr20} and the first two approaches cannot be directly applied.
Similar to our work, Yang~\etal~\cite{yang:iccv21} enforce the consistency for two different views and modalities in hand keypoint regression.
Mean teacher~\cite{tarvainen:iclr17} provides teacher-student training with consistency regularization, which regularizes a teacher network by a student's weights and avoids over-fitting to incorrect predictions.
Unlike~\cite{yang:iccv21}, we propose to integrate confidence estimation into the consistency training and adopt the teacher-student scheme with two networks.
To encourage the two networks to have different representations, we propose a distillation-based update rule instead of updating the teacher with the exponential moving average~\cite{tarvainen:iclr17}.

\section{Proposed Method}\label{sec:proposal}
\begin{figure*}[t]
\centering
\includegraphics[width=1\hsize]{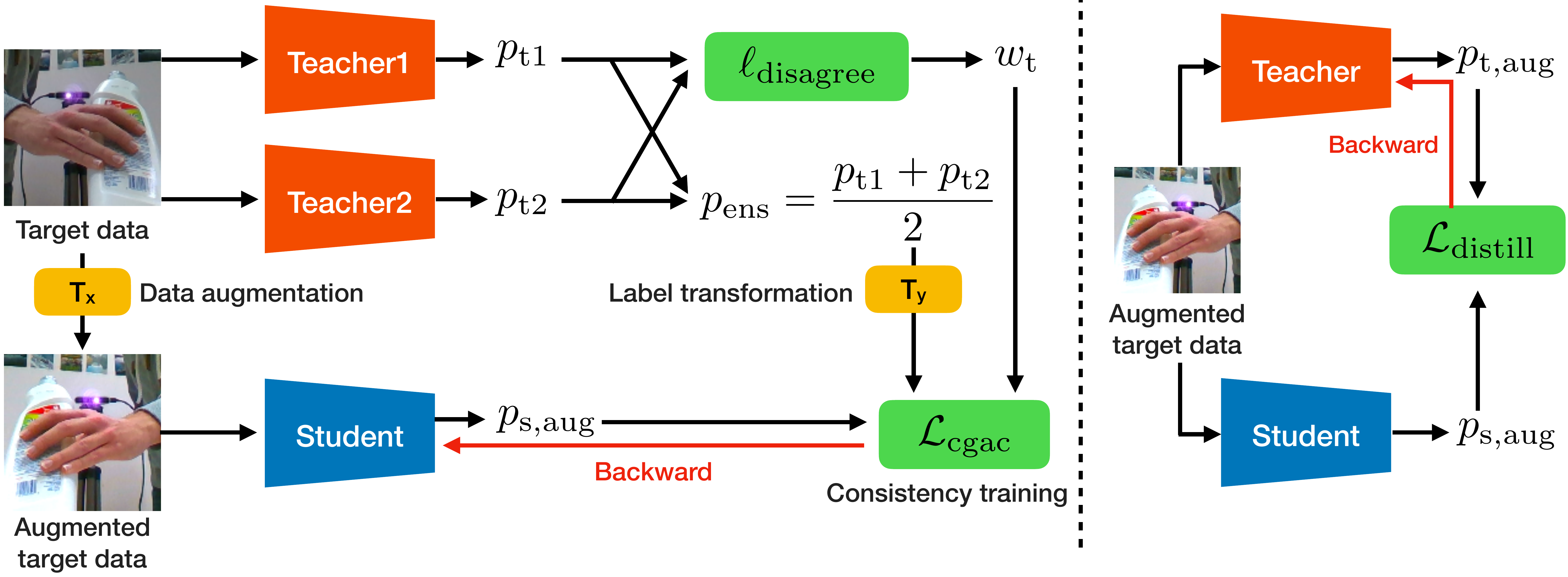}
\caption{\tbf{Method overview}.
\tbf{Left}: Student training with confidence-aware geometric augmentation consistency. 
The student learns from the consistency between its prediction and the two teachers' predictions. The training is weighted by the target confidence computed by the divergence of both teachers.
\tbf{Right}: Teacher training with knowledge distillation. Each teacher independently learns to match the student's predictions.
The task index $k$ is omitted for simplicity.
}
\label{fig:method}
\end{figure*}

In this section, we present our proposed self-training domain adaptation with confidence estimation for adapting hand keypoint regression and hand segmentation.
We first present our problem formulation and network initialization with supervised learning from source data. We then introduce our proposed modules: (1) geometric augmentation consistency, (2) confidence weighting by using two networks, and (3) teacher-student update via knowledge distillation.
As shown in Fig.~\ref{fig:method}, our adaptation is done with two different networks (teachers) for confidence estimation and another network (student) for self-training of both tasks.

\tbf{Problem formulation.}
Given labeled images from one source domain and unlabeled images from another target domain, we aim to jointly estimate hand keypoint coordinates and pixel-level hand masks on the target domain. 
We have a source image $\bs{x}_{\mathrm{s}}$ drawn from a set $X_{\mathrm{s}} \subset \mathbb{R}^{H \times W \times 3}$, its corresponding labels $(\bs{y}_{\mathrm{s}}^{\mathrm{p}}, \bs{y}_{\mathrm{s}}^{\mathrm{m}})$, and a target image $\bs{x}_{\mathrm{t}}$ drawn from a set $X_{\mathrm{t}} \subset \mathbb{R}^{H \times W \times 3}$.
The pose label $\bs{y}_{\mathrm{s}}^{\mathrm{p}}$ consists of the 2D keypoint coordinates of 21 hand joints obtained from a set $Y_{\mathrm{s}}^{\mathrm{p}} \subset \mathbb{R}^{21 \times 2}$, while the mask label $\bs{y}_{\mathrm{s}}^{\mathrm{m}}$ denotes a binary mask obtained from $Y_{\mathrm{s}}^{\mathrm{m}} \subset {(0, 1)}^{H \times W}$.
A network parameterized by $\bs{\theta}$ learns the mappings $f^{k}(x;\bs{\theta}): X \to Y^{k}$ where $k \in \{\mathrm{p}, \mathrm{m}\}$ represents the indicator for both tasks. 

\tbf{Initialization with supervised learning.}
To initialize networks used in our adaptation, we train the network $f$ on the labeled source data following multi-task learning.
Given the labeled dataset $(X_{\mathrm{s}}, Y_{\mathrm{s}})$ and the network $\bs{\theta}$, a supervised loss function is defined as 
\begin{align}
\mathcal{L}_{\mathrm{task}} \left(\bs{\theta}, X_{\mathrm{s}}, Y_{\mathrm{s}} \right) = \sum_{k} \lambda^{k} \mathbb{E}_{(\bs{x}_{\mathrm{s}}, \bs{y}_{\mathrm{s}}^{k}) \sim (X_{\mathrm{s}}, Y_{\mathrm{s}}^{k})} \left[ \mathcal{L}^{k} (\bs{p}_s^{k}, \bs{y}_{\mathrm{s}}^{k}) \right],
\label{eq:task}
\end{align}
where $Y_{\mathrm{s}} = \{Y_{\mathrm{s}}^{\mathrm{p}}, Y_{\mathrm{s}}^{\mathrm{m}}\}$ and $\bs{p}_{\mathrm{s}}^{k}=f^{k}(\bs{x}_{\mathrm{s}}; \bs{\theta})$.
$\mathcal{L}^{k}(\cdot, \cdot): Y^{k} \times Y^{k} \to \mathbb{R}^{+}$ is a loss function in each task and $\lambda^{k}$ is a hyperparameter to balance the two tasks.
We use a smooth L1 loss~\cite{huang:aaai20,ren:bmvc19} as $\mathcal{L}^{\mathrm{p}}$ and a binary cross-entropy loss as $\mathcal{L}^{\mathrm{m}}$.

\subsection{Geometric Augmentation Consistency}
Inspired by semi-supervised learning using hand keypoint consistency~\cite{yang:iccv21}, we advance a unified training with consistency for both hand keypoint regression and hand segmentation.
We expect that joint adaption of both tasks will allow one task to provide useful cues to the other task in consistency training, as studied in supervised or weakly-supervised learning setups~\cite{wang:tcsvt19,zhang:icassp20,goudie:fg17,neverova:cviu17,chen:access18}.
We design consistency training by predicting the location of hand keypoints and hand pixels in a given geometrically transformed image, including rotation and transition.
This consistency under geometric augmentation encourages the network to learn against positional bias in the target domain, which helps capture the hand structure related to poses and regions.
Specifically, given a paired augmentation function $(T_{\mathrm{x}}, T_{\mathrm{y}}^{k}) \sim \mathcal{T}$ for an image and an label, we 
generate the prediction on the target images $\bs{p}_{\mathrm{t}}^{k}=f^{k} \left(\bs{x}_{\mathrm{t}}; \bs{\theta}\right)$ and the augmented target images  $\bs{p}_{\mathrm{t,aug}}^{k}=f^{k} \left(T_{\mathrm{x}}(\bs{x}_{\mathrm{t}}); \bs{\theta}\right)$. 
We define the loss function of geometric augmentation consistency (GAC) $\mathcal{L}_{\mathrm{gac}}$ between $\bs{p}_{\mathrm{t,aug}}^{k}$ and $T_{\mathrm{y}}^{k}(\bs{p}_{\mathrm{t}})$ as
\begin{align}
\mathcal{L}_{\mathrm{gac}} \left(\bs{\theta}, {X}_t, \mathcal{T} \right) = \mathbb{E}_{\bs{x}_{\mathrm{t}}, (T_{\mathrm{x}}, T_{\mathrm{y}}^{\mathrm{p}}, T_{\mathrm{y}}^{\mathrm{m}})} \left[ \sum_{k \in \{\mathrm{p}, \mathrm{m}\}} \tilde{\lambda}^{k} \mathcal{\tilde{L}}^{k} \left(\bs{p}_{\mathrm{t,aug}}^{k}, T_{\mathrm{y}}^{k}(\bs{p}_{\mathrm{t}}^{k}) \right) \right].
\label{eq:gac}
\end{align}
To correct the augmented prediction $\bs{p}_{\mathrm{t,aug}}^{k}$ by $T_{\mathrm{y}}^{k}(\bs{p}_{\mathrm{t}})$, we stop the gradient update for $\bs{p}_{\mathrm{t}}^{k}$, which can be viewed as the supervision to $\bs{p}_{\mathrm{t,aug}}^{k}$.
We use the smooth L1 loss (see Equation~\ref{eq:task}) as $\mathcal{\tilde{L}}^{\mathrm{p}}$ and a mean squared error as $\mathcal{\tilde{L}}^{\mathrm{m}}$.
We introduce $\tilde{\lambda}^{k}$ as a hyperparameter to control the balance of the two tasks.
The augmentation set $\mathcal{T}$ contains the geometric augmentation and photometric augmentation, such as color jitter and blurring.
We set $T_{\mathrm{y}}(\cdot)$ to align geometric information to the augmented input $T_{\mathrm{x}}(\bs{x}_{\mathrm{t}})$. 
For example, we apply rotation $T_{\mathrm{y}}(\cdot)$ to the outputs $\bs{p}_{\mathrm{t}}^{k}$ with the same degree of rotation $T_{\mathrm{x}}(\cdot)$ to the input $\bs{x}_{\mathrm{t}}$.

\subsection{Confidence Estimation by Two Separate Networks}\label{sec:cofidence}
Since the target predictions are not always reliable, we aim to incorporate the estimated confidence weight for each target instance into the consistency training.
In Equation~\ref{eq:gac}, the generated outputs $\bs{p}_{\mathrm{t}}^{k}$ that is the supervision to $\bs{p}_{\mathrm{t,aug}}^{k}$ may be unstable and noisy due to the domain gap between source and target domains.
Due to that, the network trained with the consistency readily overfits to the incorrect supervision $\bs{p}_{\mathrm{t}}^{k}$, which is known as confirmation bias~\cite{arazo:ijcnn20,tarvainen:iclr17}.
To reduce the bias, it is crucial to assign a low importance (confidence) weight to the consistency training with the incorrect supervision.
This enables the network to learn primarily from reliable supervision while avoiding being biased to such erroneous predictions.
In classification tasks, predicted class probability can serve as the confidence, while these measures are not trivially defined and available in regression tasks.
To estimate the confidence of keypoint predictions, Yang~\etal~\cite{yang:iccv21} measure the confidence of 3D hand keypoints by the distance to the fitted 3D hand template, but the hand template fitting is an ill-posed problem for 2D hands and is not applicable to hand segmentation.
Dropout~\cite{gal:icml16,cai:cvpr20,cai:arxiv21} is a generic way of estimating uncertainty (confidence), calculated by the variance of multiple stochastic forwards.
However, the estimated confidence is biased to the current state of the training network because the training and confidence estimation are done by a single network.
When the training network works poorly, the confidence estimation becomes readily unreliable.

To perform reliable confidence estimation for both tasks, we propose a confidence measure by computing the divergence of two predictions.
Specifically, we introduce two networks (\aka, teachers) for the confidence estimation and the estimated confidence is used to train another network (\aka, student) for the consistency training.
The architecture of the teachers is identical, yet they have different learning parameters.
We observe that when the divergence of the two predictions from the teachers for a target instance is high, the predictions of both networks become unstable.
In contrast, a lower divergence indicates that the two teacher networks predict stably and agree on their predictions.
Thus, we use the divergence for representing the target confidence.
Given the teachers $\bs{\theta}^{\mathrm{tch1}}, \bs{\theta}^{\mathrm{tch2}}$, we define a disagreement measure $\ell_{\mathrm{disagree}}$ to compute the divergence as
\begin{align}
 \ell_{\mathrm{disagree}} \left(\bs{\theta}^{\mathrm{tch1}}, \bs{\theta}^{\mathrm{tch2}}, \bs{x}_{\mathrm{t}}\right) = \sum_{k \in \{\mathrm{p}, \mathrm{m}\}} \tilde{\lambda}^{k} \mathcal{\tilde{L}}^{k} (\bs{p}_{\mathrm{t1}}^{k}, \bs{p}_{\mathrm{t2}}^{k}),
\label{eq:disagree}
\end{align}
where $\bs{p}_{\mathrm{t1}}^{k}=f^{k}(\bs{x}_{\mathrm{t}}; \bs{\theta}^{\mathrm{tch1}})$ and $\bs{p}_{\mathrm{t2}}^{k}=f^{k}(\bs{x}_{\mathrm{t}}; \bs{\theta}^{\mathrm{tch2}})$.

\begin{figure*}[t!]
\centering
\includegraphics[width=0.9\hsize]{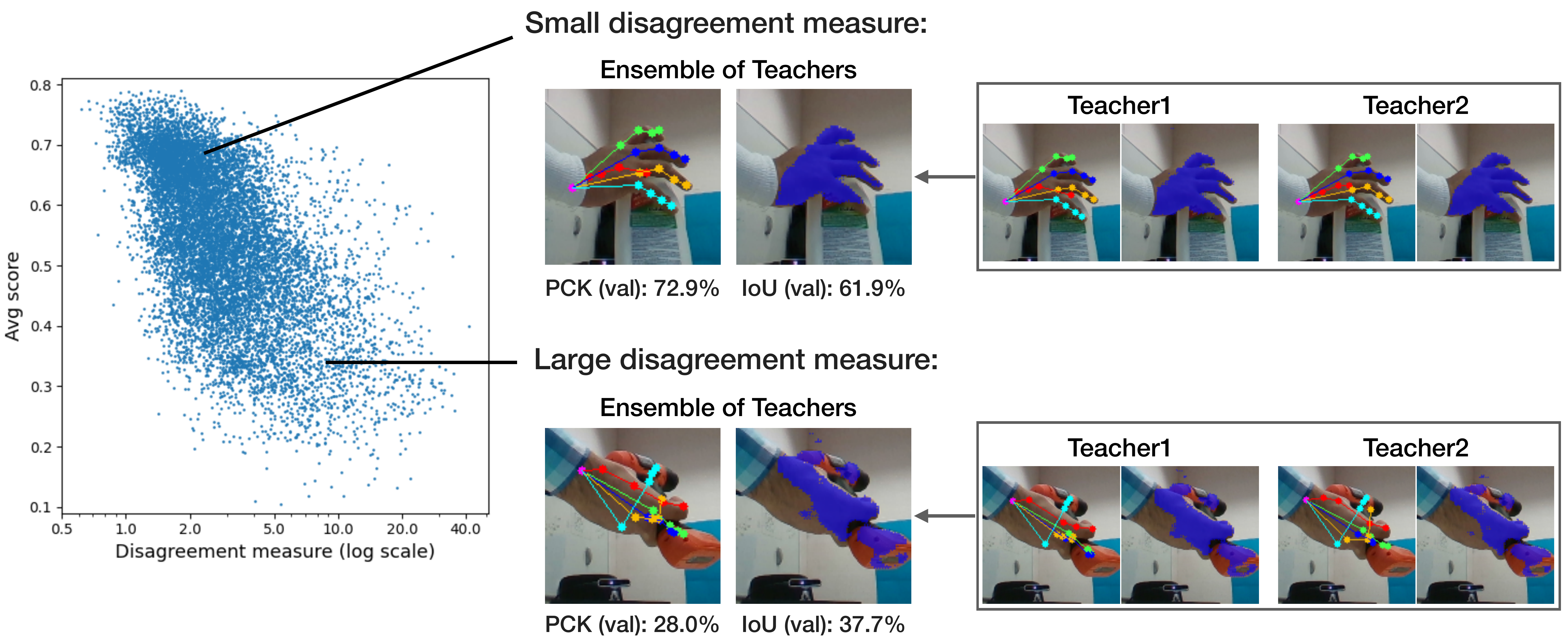}
\caption{\tbf{The correlation between a disagreement measure and task scores.} Target instances with smaller disagreement values between the two teacher networks tend to have higher task scores.
}
\label{fig:disagree}
\end{figure*}

As a proof of concept, we visualize the correlation between the disagreement measure and a validation score averaged over evaluation metrics of the two tasks (PCK and IoU) in Fig.~\ref{fig:disagree}.
We compute the score between the ensemble of the teachers' predictions $\bs{p}_{\mathrm{ens}}^{k}=\left(\bs{p}_{\mathrm{t1}}^{k} + \bs{p}_{\mathrm{t2}}^{k}\right)/2$ and its ground truth in the validation set on HO3D~\cite{hampali:cvpr20}.
The instances with a small disagreement measure tend to have high validation scores. 
In contrast, the instances with a high disagreement measure entail false predictions, \eg, detecting the hand-held object as a hand joint and hand class.
When the disagreement measure was high at the bottom of Fig.~\ref{fig:disagree}, we found that both predictions were particularly unstable on the keypoints of the ring finger (yellow).
This study shows that the disagreement measure can represent the correctness of the target predictions.

With the disagreement measure $\ell_{\mathrm{disagree}}$, we define a confidence weight $w_{\mathrm{t}} \in [0, 1]$ for assigning importance to the consistency training.
We compute the weight $w_{\mathrm{t}}$ as $w_{\mathrm{t}} = 2 \left( 1 - \mathrm{sigm}\left(\lambda_{\mathrm{d}} \ell_{\mathrm{disagree}} \left(\bs{\theta}^{\mathrm{tch1}}, \bs{\theta}^{\mathrm{tch2}}, \bs{x}_{\mathrm{t}}\right)\right)\right)$ where $w_{\mathrm{t}}$ is a normalized disagreement measure with sign inversion, $\mathrm{sigm}(\cdot)$ denotes a sigmoid function, and $\lambda_{\mathrm{d}}$ controls the scale of the measure.
With the confidence weight $w_{\mathrm{t}}$, we enforce the consistency training between the student's prediction on the augmented target images $\bs{p}_{\mathrm{s,aug}}^{k}$ and the ensemble of the two teachers' predictions $\bs{p}_{\mathrm{ens}}^{k}$.
Our proposed loss function of confidence-aware geometric augmentation consistency (C-GAC)
$\mathcal{L}_{\mathrm{cgac}}$ for the student  $\bs{\theta}^{\mathrm{stu}}$ is formulated as
\begin{align}
\mathcal{L}_{\mathrm{cgac}} \left(\bs{\theta}^{\mathrm{stu}}, \bs{\theta}^{\mathrm{tch1}}, \bs{\theta}^{\mathrm{tch2}}, {X}_{\mathrm{t}}, \mathcal{T} \right) &= \mathbb{E}_{\bs{x}_{\mathrm{t}}, (T_{\mathrm{x}}, T_{\mathrm{y}}^{\mathrm{p}}, T_{\mathrm{y}}^{\mathrm{m}})} \left[w_{\mathrm{t}} \sum_{k \in \{\mathrm{p}, \mathrm{m}\}} \tilde{\lambda}^{k} \mathcal{\tilde{L}}^{k} \left(\bs{p}_{\mathrm{s,aug}}^{k}, T_{\mathrm{y}}^{k}(\bs{p}_{\mathrm{ens}}^{k}) \right) \right],
\label{eq:cgac}
\end{align}
where $\bs{p}_{\mathrm{s,aug}}^{k}=f^{k} \left(T_{\mathrm{x}}(\bs{x}_{\mathrm{t}}); \bs{\theta}^{\mathrm{stu}}\right)$.
Following~\cite{tarvainen:iclr17,pham:cvpr21}, we design the student prediction $\bs{p}_{\mathrm{s,aug}}^{k}$ to be supervised by the teachers.
We generate the teachers' prediction by doing ensemble $\bs{p}_{\mathrm{ens}}^{k}$, which is better than the prediction of either teacher.

\subsection{Teacher-Student Update by Knowledge Distillation}\label{sec:distill}
In addition to the student's training, we formulate an update rule for the two teacher networks by using knowledge distillation.
Since $\ell_{\mathrm{disagree}}$ would not work if the two teachers had the same output values, we aim to learn two teachers that have different representations yet keep high task performance as co-training works~\cite{blum:colt98,qiao:eccv18,chen:nips11,saito:icml17}.
In a prior teacher-student update, Tarvainen~\etal~\cite{tarvainen:iclr17} found that the teacher's update by an exponential moving average (EMA), which averages the student's weights iteratively, makes the teacher's learning more slowly and mitigates the confirmation bias as discussed in Section~\ref{sec:cofidence}.
While this EMA-based teacher-student framework is widely used in various domain adaptation tasks~\cite{french:iclr18,cai:cvpr19,li:arxiv21,yan:lgrs22,ge:iclr20}, naively applying the EMA rule to the two teachers would produce exactly the same weights for both networks.

To prevent this, we propose independent knowledge distillation for building two different teachers.
The distillation matches the teacher-student predictions in the output level. 
To let both networks have different parameters, we train the teachers from different mini-batches and using stochastic augmentation as
\begin{align}
  \mathcal{L}_{\mathrm{distill}} \left(\bs{\theta}, \bs{\theta}^{\mathrm{stu}}, X_{\mathrm{t}},  \mathcal{T} \right) =  \mathbb{E}_{\bs{x}_{\mathrm{t}},T_{\mathrm{x}}} \left[ \sum_{k \in \{\mathrm{p}, \mathrm{m}\}} \tilde{\lambda}^{k} \mathcal{\tilde{L}}^{k} (\bs{p}_{\mathrm{t,aug}}^{k}, \bs{p}_{\mathrm{s,aug}}^{k}) \right],
\label{eq:distill}
\end{align}
where $~\bs{\theta} \in \{\bs{\theta}^{\mathrm{tch1}}, \bs{\theta}^{\mathrm{tch2}}\}$, $\bs{p}_{\mathrm{t,aug}}^{k}=f^{k} \left(T_{\mathrm{x}}(\bs{x}_{\mathrm{t}}); \bs{\theta}\right)$, and  $\bs{p}_{\mathrm{s,aug}}^{k}=f^{k}(T_{\mathrm{x}}\left(\bs{x}_{\mathrm{t}}\right); \bs{\theta}^{\mathrm{stu}})$.
The distillation loss $ \mathcal{L}_{\mathrm{distill}}$ is used for updating the teacher networks only.
This helps the teachers to adapt to the target domain more carefully than the student and avoid falling into exactly the same predictions on a target instance. 

\subsection{Overall Objectives}\label{sec:overall}
Overall, the objective of the student's training consists of the supervised loss (Equation~\ref{eq:task}) from the source domain and the self-training with confidence-aware geometric augmentation consistency (Equation~\ref{eq:cgac}) in the target domain as
\begin{align}
\min_{\bs{\theta}^{\mathrm{stu}}}  \mathcal{L}_{\mathrm{task}} \left(\bs{\theta}^{\mathrm{stu}}, X_{\mathrm{s}}, Y_{\mathrm{s}} \right) +  \mathcal{L}_{\mathrm{cgac}} \left(\bs{\theta}^{\mathrm{stu}}, \bs{\theta}^{\mathrm{tch1}}, \bs{\theta}^{\mathrm{tch2}}, {X}_t, \mathcal{T} \right).
\label{eq:loss_stu}
\end{align}
The two teachers are asynchronously trained with the distillation loss (Equation~\ref{eq:distill}) in the target domain, which is formulated as
\begin{align}
\min_{\bs{\theta}}  \mathcal{L}_{\mathrm{distill}} \left(\bs{\theta}, \bs{\theta}^{\mathrm{stu}}, X_{\mathrm{t}},  \mathcal{T} \right),
\label{eq:loss_tch}
\end{align}
where $\bs{\theta} \in \{\bs{\theta}^{\mathrm{tch1}}, \bs{\theta}^{\mathrm{tch2}}\}$.
Since the teachers are updated carefully and can perform better than the student, we use the ensemble of the two teachers' predictions for a final output in inference.

\section{Experiments}\label{sec:experiment}
In this section, we first present our experimental datasets and implementation details and then provide quantitative and qualitative results along with the ablation studies. We analyze our proposed method by comparing it with several existing methods in three different domain adaptation settings. We also show qualitative results by applying our method to in-the-wild egocentric videos.

\subsection{Experiment Setup}\label{sec:exp_setup}
\tbf{Datasets.}
We experimented with several hand datasets including a variety of hand-object interactions, the annotation of 2D hand keypoints, and hand masks as follows. 
We adopted \tbf{DexYCB~\cite{chao:cvpr21}} dataset as our source dataset since it contains a large amount of training images, their corresponding labels, and natural hand-object interactions.
We chose to use the following datasets as our target datasets: \tbf{HO3D~\cite{hampali:cvpr20}} captured in different environments with the same YCB objects~\cite{berk:ram15} as the source dataset, \tbf{HanCo~\cite{zimmermann:arxiv21}} captured in a multi-camera studio and generated with synthesized backgrounds, and \tbf{FPHA~\cite{hernando:cvpr18}} captured by a first-person view.
We also used \tbf{Ego4D~\cite{grauman:cvpr22}} to verify the effectiveness of our method in real-world scenarios.
During training, we used cropped images of the hand regions from the original images as input.

\tbf{Implementation details.}
Our teacher-student networks share an identical network architecture, which consists of a unified feature extractor and task-specific branches for hand keypoint regression and hand segmentation. 
For training our student network, we used the Adam optimizer~\cite{kingma:iclr14} with a learning rate of $10^{-5}$, while the learning rate of the teacher networks was set to $5 \times 10^{-6}$.
We set the hyperparameters ($\lambda^{\mathrm{p}}(=\tilde{\lambda}^{\mathrm{p}})$, $\lambda^{\mathrm{m}}$, $\tilde{\lambda}^{\mathrm{m}}$, $\lambda_{\mathrm{d}}$) to $(10^{7}, 10^{2}, 5, 0.5)$.
Since both task-specific branches have different training speeds, we began our adaptation with the backbone and keypoint regression branch. 
We then trained all sub-networks, including the hand segmentation branch.
We report the percentage of correct keypoints (PCK) and the mean joint position error (MPE) for hand keypoint regression, and the intersection over union (IoU) for hand segmentation. 

\tbf{Baseline methods.} We compared quantitative performance with the following methods. \tbf{Source only} denotes the network trained on the source dataset without any adaptation. 
To compare with another adaptation approach with adversarial training, we trained \tbf{DANN~\cite{ganin:icml15}} that aligns marginal feature distributions between domains, and \tbf{RegDA~\cite{jiang:cvpr21}} with an adversarial regressor that optimizes domain disparity.
In addition, we implemented several self-training adaptation methods by replacing pseudo-labeling with the consistency training.
\tbf{GAC} is a simple baseline with the consistency training updated by Equation~\ref{eq:gac}.
\tbf{GAC + UMA~\cite{cai:cvpr20}} is a GAC method with confidence estimation by Dropout~\cite{gal:icml16}.
\tbf{GAC + CPL~\cite{ohkawa:access21}} is a GAC method with confident instance selection using the agreement with another network.
\tbf{GAC + MT~\cite{tarvainen:iclr17}} is a GAC method with the single-teacher-single-student architecture using EMA for the teacher update.
\tbf{Target only} indicates the network trained on the target dataset with labels, which shows an empirical performance upper bound.

\tbf{Our method.}
We denote our full method as \tbf{C-GAC} introduced in Section~\ref{sec:overall}.
As an ablation study, we present a variant of the proposed method as \tbf{GAC-Distill} with a teacher-student pair, which is updated by the consistency training (Equation~\ref{eq:gac}) and the distillation loss (Equation~\ref{eq:distill}).
\tbf{GAC-Distill} is different from \tbf{GAC + MT} only in the way of the teacher update.

\subsection{Quantitative Results}
We show the results of three adaptation settings: DexYCB → \{HO3D, HanCo, FPHA\} in Tables~\ref{tbl:d2ho} and~\ref{tbl:d2haf}. 
We then provide detailed comparisons of our method.

\tabledho
\tbf{DexYCB → HO3D.}
Table~\ref{tbl:d2ho} shows the results of the adaptation from DexYCB to HO3D where the grasping objects are overlapped.
The baseline of the consistency training (\tbf{GAC}) was effective in learning target images in both tasks.
Our proposed method (\tbf{C-GAC}) improved by $5.3/14.4$ in the average task score from the source-only performance.
The method also outperformed all comparison methods and achieved close performance to the upper bound.

\tabledhaf

\tbf{DexYCB → HanCo.}
Table~\ref{tbl:d2haf} shows the results of the adaptation from DexYCB to HanCo across laboratory setups.
The source-only network less generalized to the target domain because the HanCo has diverse backgrounds, while \tbf{GAC} succeeded in adapting up to $47.4/47.9$ in the average score.
Our method \tbf{C-GAC} showed further improved results in hand keypoint regression.

\tbf{DexYCB → FPHA.}
Table~\ref{tbl:d2haf} also shows the results of the adaptation from DexYCB to FPHA, which captures egocentric users' activities.
Since hand markers and in-the-wild target environments cause large appearance gaps, the source-only performance performed the most poorly among the three adaption settings. 
In this challenging setting, \tbf{RegDA} and \tbf{GAC + UMA} performed well for hand segmentation, while their performance on hand keypoint regression was inferior to the \tbf{GAC} baseline.
Our method \tbf{C-GAC} further improved than the \tbf{GAC} method in the MPE and IoU metrics and exhibited stability in adaptation training among the comparison methods.

\tbf{Comparison to different confidence estimation methods.}
We compare the results with existing confidence estimation methods.
\tbf{GAC + UMA} and \tbf{GAC + CPL} estimate the confidence of target predictions by computing the variance of multiple stochastic forwards and the task scores between a training network and an auxiliary network, respectively. 
\tbf{GAC + UMA} performed effectively on DexYCB → FPHA, whereas the performance gain was thin in the other settings compared to \tbf{GAC}. 
\tbf{GAC + CPL} worked well for keypoint regression on DexYCB → HO3D, but it cannot address the other settings with a large domain gap well since the prediction of the auxiliary network became unstable.
Although these prior methods had different disadvantages depending on the settings, our method \tbf{C-GAC} using the divergence of the two teachers for confidence estimation performed stably in the three settings.

\tbf{Comparison to standard teacher-student update.}
We compare our teacher update with the update with an exponential moving average (EMA)~\cite{tarvainen:iclr17}.
The EMA-based update (\tbf{GAC-MT}) degraded the performance from the source only in hand segmentation in Table~\ref{tbl:d2ho}. 
This suggests that the EMA update can be sensitive to the task.
In contrast, our method \tbf{GAC-Distill} matching the teacher-student predictions in the output level did not produce such performance degeneration and worked more stably.

\tbf{Comparison to adversarial adaptation methods.}
We compared our method with another major adaptation approach with adversarial training.
In Tables~\ref{tbl:d2ho} and~\ref{tbl:d2haf}, the performance of \tbf{DANN} and \tbf{RegDA} was mostly worse than the consistency-based baseline \tbf{GAC}.
We found that instead of matching features between both domains~\cite{ganin:icml15,jiang:cvpr21}, directly learning target images by the consistency training was critical in the adaptation of our tasks.

\tbf{Comparison to an off-the-shelf hand pose estimator.}
We tested the generalization ability of an open-source library for pose estimation: \tbf{OpenPose}~\cite{openpose}.
It resulted in $15.75 / 12.72$, $18.31 / 18.42$, and $29.02$ in the MPE on HO3D, HanCo, and FPHA, respectively.
Since it is built on multiple source datasets~\cite{joo:cvpr15,andriluka:cvpr14,nzsl}, the baseline showed higher generalization than the source-only network.
However, the performance did not exceed our proposed method in the MPE.
This shows that generalizing hand keypoint regression to other datasets is still challenging, and our adaptation framework supports improving target performance.

\begin{figure*}[t]
\centering
\includegraphics[width=1\hsize]{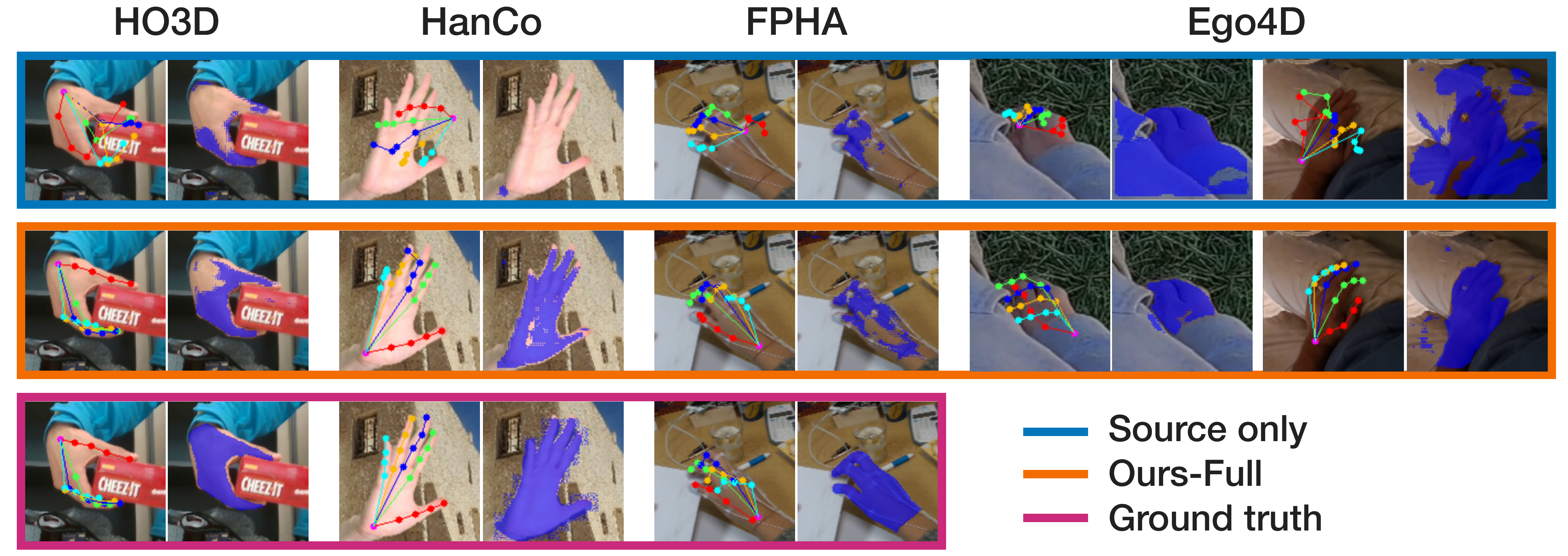}
\caption{\tbf{Qualitative results.}
We show qualitative examples of the source-only network (top), the Ours-Full method (middle), and ground truth (bottom) on HO3D~\cite{hampali:cvpr20}, HanCo~\cite{zimmermann:arxiv21}, FPHA~\cite{hernando:cvpr18}, and Ego4D~\cite{grauman:cvpr22} without ground truth.
}
\label{fig:qual}
\end{figure*}

\subsection{Qualitative Results}
We show the qualitative results of hand keypoint regression and hand segmentation in Fig.~\ref{fig:qual}. 
When hands are occluded in HO3D and FPHA or the backgrounds are diverse in HanCo, the keypoint prediction of the source only (top) represented infeasible hand poses and hand segmentation was too noisy or missing.
However, our method \tbf{C-GAC} (middle) corrected the hand keypoint errors and improved to localize hand regions.
Hand segmentation in FPHA was still noisy because visible white markers obstructed hand appearance.
We can also see distinct improvements in the Ego4D dataset.
We provide additional qualitative analysis in adaptation to the Ego4D beyond countries, cultures, ages, indoors/outdoors, and performing tasks with hands in our supplementary material.

\subsection{Ablation Studies}
\tbf{Effect of confidence estimation.}
To confirm the effect of our proposed confidence estimation, we compare our full method
\tbf{C-GAC} and our ablation model \tbf{GAC-Distill} without the confidence weighting.
In Tables~\ref{tbl:d2ho} and~\ref{tbl:d2haf}, while \tbf{GAC-Distill} mostly surpassed the comparison methods in most cases, \tbf{C-GAC} showed further performance gain in all three adaptation settings.

\tbf{Multi-task \vs single-task adaptation.}
We studied the effect of our multi-task adaptation compared with single-task adaptation on DexYCB → HO3D.
The single-task adaptation results are $50.1/51.0$ in the PCK and $58.2/57.7$ in the IoU.
Compared to Table~\ref{tbl:d2ho}, our method in the multi-task setting improved by $2.7 / 2.6$ over the single-task adaption in hand segmentation while it provided marginal gain in hand keypoint regression.
This shows that the adaptation of hand keypoint regression helps to localize hand regions in the target domain.

\begin{figure*}[t]
\centering
\includegraphics[width=0.95\hsize]{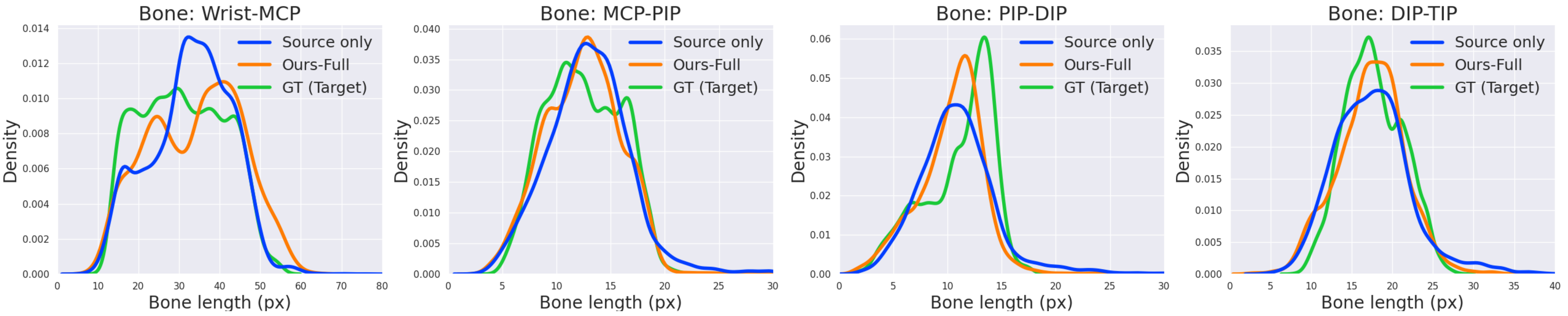}
\caption{\tbf{Visualization of bone length distributions.}
We show the distributions of the bone length between hand joints, namely, Wrist,  metacarpophalangeal (MCP), proximal interphalangeal (PIP), distal interphalangeal (DIP), and fingertip (TIP).
Using kernel density estimation, we plotted the density of the bone length for the predictions of the source only, the Ours-Full method, and ground truth on test data of HO3D~\cite{hampali:cvpr20}.
}
\label{fig:bone}
\end{figure*}

\tbf{Bone length distributions.}
To study our adaptation results in each hand joint, we show the distributions of bone length between hand joints in Fig.~\ref{fig:bone}.
In Wrist-MCP, PIP-DIP, and DIP-TIP, the distribution of the source-only prediction on target images (blue) was far from that of the target ground truth (green), whereas our method (orange) improved to approximate the target distribution (green).
In MCP-PIP, we could not observe such clear differences because the source-only model already represented the target distribution well.
This indicates that our method improved to learn hand structure near the palm and fingertips.

\section{Conclusion}\label{sec:conclusion}
In this work, we tackled the problem of joint domain adaptation of hand keypoint regression and hand segmentation. 
Our proposed method consists of the self-training with geometric augmentation consistency, confidence weighting by the two teacher networks, and the teacher-student update by knowledge distillation.
The consistency training under geometric augmentation served to learn the unlabeled target images for both tasks.
The divergence of the predictions from two teacher networks could represent the confidence of each target instance, which enables the student network to learn from reliable target predictions.
The distillation-based teacher-student update guided the teachers to learn from the student carefully and mitigated over-fitting to the noisy predictions.
Our method delivered state-of-the-art performance on the three adaptation setups.
It also showed improved qualitative results in the real-world egocentric videos.

\subsection*{Acknowledgments}
This work was supported by JST ACT-X Grant Number JPMJAX2007, JSPS Research Fellowships for Young Scientists, JST AIP Acceleration Research Grant Number JPMJCR20U1, and JSPS KAKENHI Grant Number JP20H04205, Japan.
This work was also supported in part by a hardware donation from Yu Darvish.

\appendix
\section{Appendix}
\subsection{Dataset Details}
\begin{itemize}
    \item \tbf{DexYCB~\cite{chao:cvpr21}} contains 582K RGB-D frames captured by 10 subjects interacting 20 different YCB objects~\cite{berk:ram15} from eight different views. In our experiment, we split the dataset by the subject IDs to create train, validation, and test sets with 212K, 71K, and 80K images, respectively.
    \item \tbf{HO3D~\cite{hampali:cvpr20}} contains 103K RGB-D frames captured by 10 subjects interacting 10 different YCB objects~\cite{berk:ram15} from a single third-person view. In our experiment, we randomly split the video sequences to train, validation, and test sets with 51K, 12K, and 8K images, respectively.
    \item \tbf{HanCo~\cite{zimmermann:arxiv21}} is an extended FreiHAND~\cite{zimmermann:iccv19} dataset captured in a multi-view camera setup with eight cameras, which consists of 518K, 106K, and 104K RGB images for training, validation, and testing, respectively. The backgrounds are randomly synthesized using diverse scenery images.
    \item \tbf{FPHA~\cite{hernando:cvpr18}} is an egocentric video dataset capturing users' actions in daily indoor environments from a first-person perspective, and their hand poses are tracked by hand magnetic sensors. It contains 69K training images and 16K validation images. Due to lacking hand mask annotation, we annotated 50 hand masks in the validation set. 
    \item \tbf{Ego4D~\cite{grauman:cvpr22}} is a collection of daily-life egocentric activity videos lasting over 3,000 hours and gathered across the world.
    Due to the lack of annotation for the two tasks, we show qualitative examples in our experiments.
    We treated each video sequence as the domain to adapt.
\end{itemize}

\subsection{Preprocessing and Augmentation}
For creating an input of a training network, we assumed to have hand center positions, cropped hand regions of the original images, and resized them to $128\times128$ pixels.
To extract hand centers and regions in Ego4D videos without ground truth, we used an off-the-shelf hand detector~\cite{shan:cvpr20}.
Inspired by~\cite{liu:iclr21,deng:cvpr21,li:arxiv21}, we used two different augmentation sets: strong augmentation for the student's learning (Equation~\red{4}) and weak augmentation for the teacher's learning (Equation~\red{5}).
We used horizontal flip, rotation, transition, gaussian blur, brightness/contrast jitter, hue/saturation/input value jitter, and cutout as the strong augmentation.
In contrast, we adopted horizontal flip, rotation, transition, and gaussian blur as the weak augmentation.

\subsection{Network Architecture and Evaluation}
For the design of our multi-task baseline model, we employed an hourglass network~\cite{newell:eccv16} as the backbone and the keypoint regression branch. 
We added 1d-convolution to its intermediate features to predict hand pixel labels.
Following hand keypoint regression methods~\cite{wei:cvpr16,newell:eccv16,ge:cvpr19}, we optimized 2D joint heatmaps for each 2D ground-truth joint location instead of joint coordinates. 

We also provide the details of out evaluation, namely, MPE, PCK, and IoU.
MPE (px) indicates the euclidean error per joint in the image coordinate.
PCK (\%) represents the percentage of joints whose MPE is smaller than a given joint error threshold, which is calculated by the area under the curve (AUC) over the joint error range [0, 20 px].
IoU (\%) measures the overlap over two masks.
We report the average score (Avg.) over PCK and IoU to evaluate multi-task performance.

\subsection{Qualitative Analysis}
\begin{figure}[H]
\centering
\includegraphics[width=1\hsize]{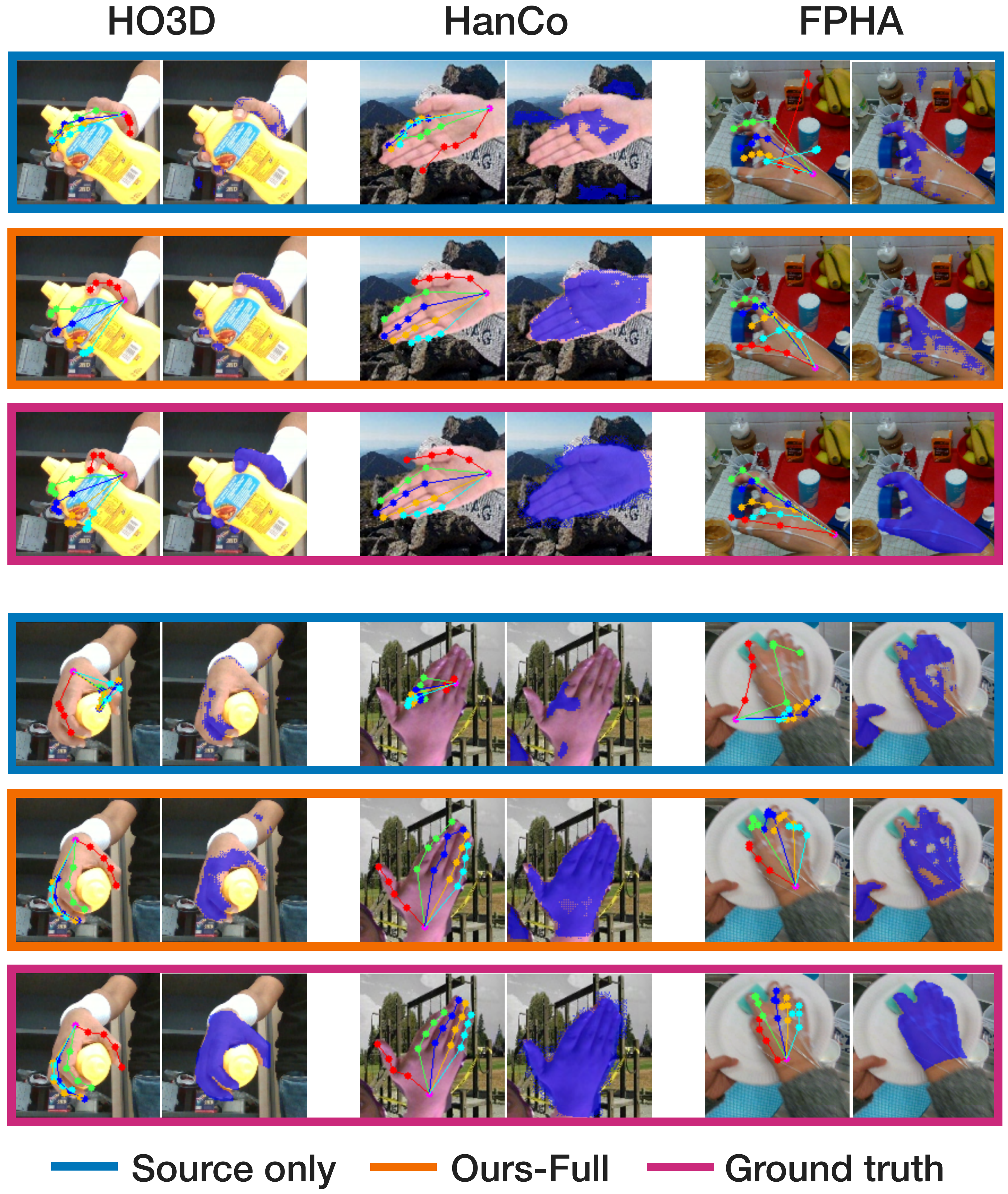}
\caption{\tbf{Additional qualitative results on HO3D~\cite{hampali:cvpr20}, HanCo~\cite{zimmermann:arxiv21}, and FPHA~\cite{hernando:cvpr18}.}
}
\label{fig:add_qual}
\end{figure}

\begin{figure}[H]
\centering
\includegraphics[width=1\hsize]{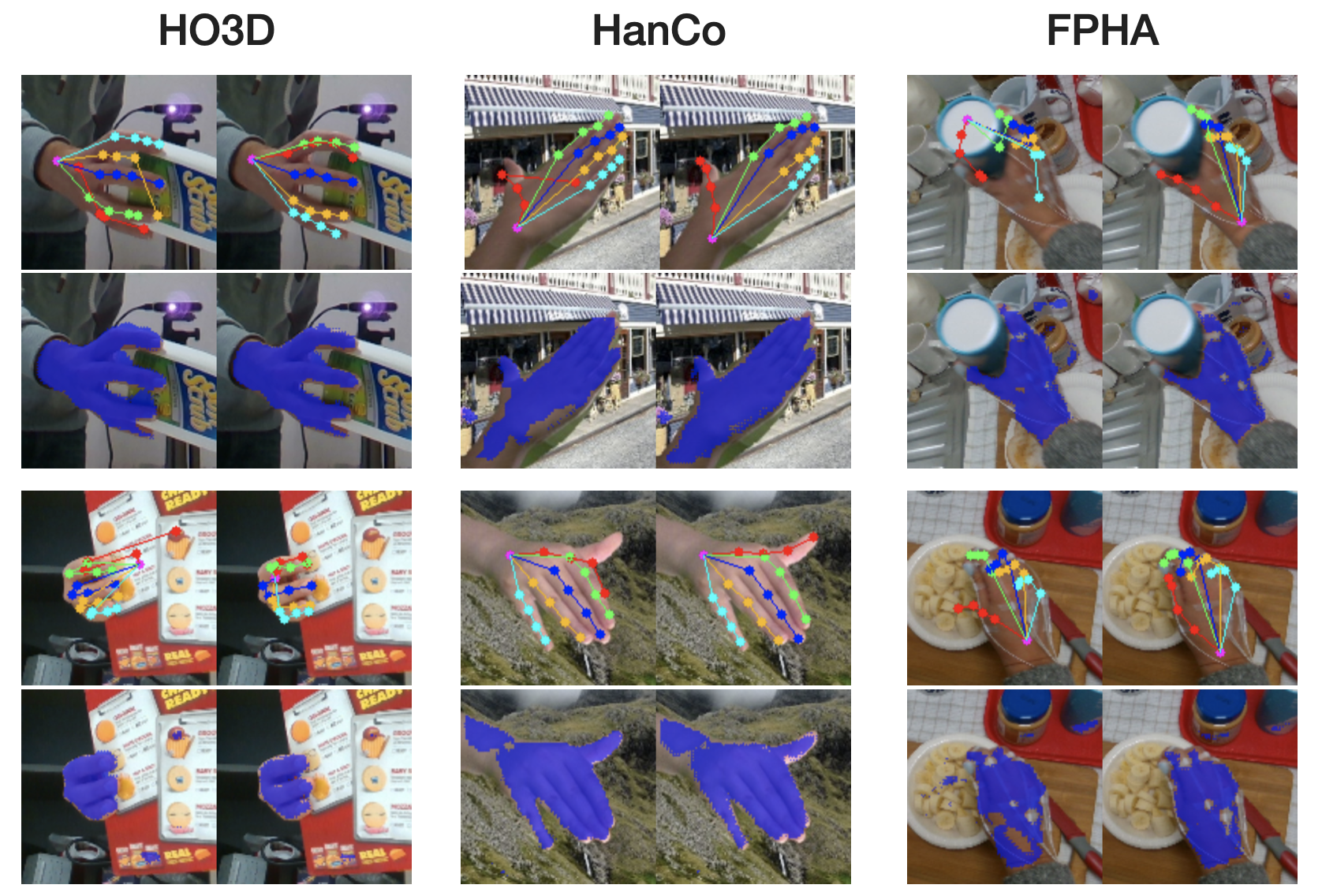}
\caption{
\tbf{Comparison between GAC and C-GAC (Ours-Full).} 
Left: GAC, Right: C-GAC (Ours-Full).
}
\label{fig:gac}
\end{figure}

\begin{figure}[H]
\centering
\includegraphics[width=1\hsize]{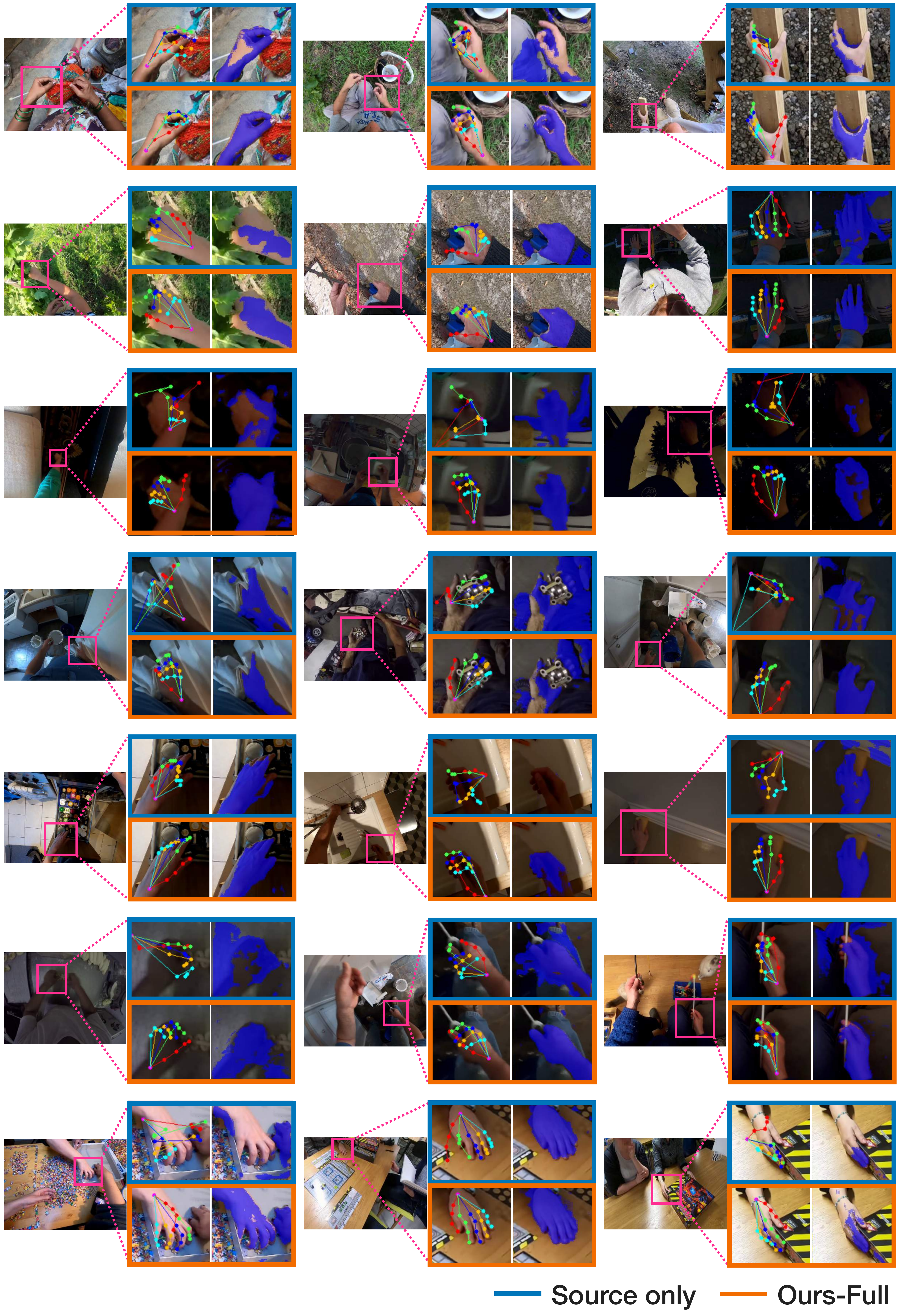}
\caption{\tbf{Additional qualitative results on Ego4D~\cite{grauman:cvpr22}.}}
\label{fig:add_ego4d}
\end{figure}

\begin{figure}[H]
\centering
\includegraphics[width=1\hsize]{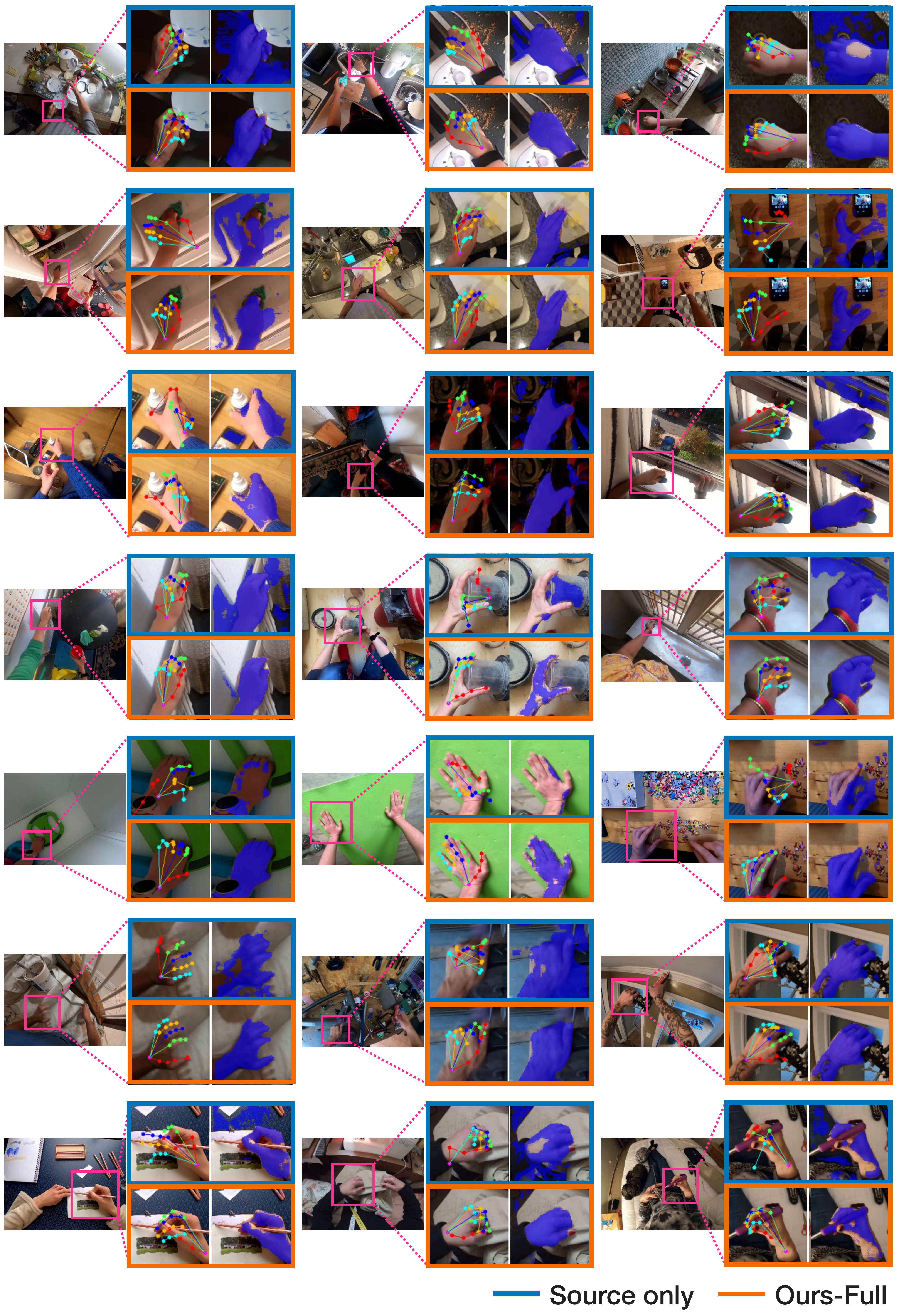}
\caption{\tbf{Additional qualitative results on Ego4D~\cite{grauman:cvpr22}.}
}
\label{fig:add_ego4d_2}
\end{figure}

In Figs.~\ref{fig:add_qual},~\ref{fig:gac},~\ref{fig:add_ego4d}, and~\ref{fig:add_ego4d_2}, we show additional qualitative results of our proposed method.
As shown in Fig.~\ref{fig:add_qual}, our method performed well when complex hand-object interactions occur on HO3D and FPHA and when the backgrounds are diverse on HanCo.
In Fig.~\ref{fig:gac}, we show qualitative comparison between GAC and C-GAC (Ours-Full).
Our full method particularly improved keypoint regression compared to the simple consistency baseline, GAC.
Our method (right) corrected the keypoint prediction of the GAC (left), which contains incorrect predictions on the position of the thumb (red).

Our method also demonstrated improved performance on Ego4D, an egocentric video dataset collected across various countries, cultures, ages, indoors/outdoors, and performing tasks with hands.
In particular, we observed that our method successfully adapted to various imaging conditions, such as outdoor environments (rows 1 and 2 in Fig.~\ref{fig:add_ego4d}), extremely dark environments (rows 3 to 6 in Fig.~\ref{fig:add_ego4d}), the second person's hands in social interactions (row 7 in Fig.~\ref{fig:add_ego4d}), \eg, playing board games, and indoor environments (Fig.~\ref{fig:add_ego4d_2}), \eg, where people perform cooking, cleaning, fitness, DIY, painting, and crafting.

\newpage 

\newpage
{\small
\bibliographystyle{ieee_fullname}
\bibliography{main.bbl}
}

\end{document}